# Using Machine Learning to Predict the Outcome of English County twenty over Cricket Matches

Stylianos Kampakis, University College London, stylianos.kampakis@gmail.com
William Thomas, University College London

## Abstract

Cricket betting is a multi-billion dollar market. Therefore, there is a strong incentive for models that can predict the outcomes of games and beat the odds provided by bookers. The aim of this study was to investigate to what degree it is possible to predict the outcome of cricket matches. The target competition was the English twenty over county cricket cup. The original features alongside engineered features gave rise to more than 500 team and player statistics. The models were optimized firstly with team features only and then both team and player features. The performance of the models was tested over individual seasons from 2009 to 2014 having been trained over previous season data in each case. The optimal model was a simple prediction method combined with complex hierarchical features and was shown to significantly outperform a gambling industry benchmark.

## 1  Introduction

As a sport cricket is played globally across 106 member states of the International Cricket Council (ICC), with an estimated 1.5 billion fans worldwide (ICC, 2012-2013). However, much of the global finance and interest is focused upon the 10 full ICC member nations and more specifically upon 'the big three' of England, Australia and India[1]. Annually the annual cricket gambling market is thought to be worth $10 billion legally and between $40-50 billion illegally, driven by the Asian markets[2]. The Western gambling markets are heavily regulated and monitored and therefore this lack of an obvious Asian betting market potentially limits the opportunity. This is perhaps the reason that limited work to date has taken place in this area.

Given the scale of the betting industry worldwide, there are obviously monetary gains for anyone with access to superior prediction techniques, whether through working with betting companies, selling predictions to professional gamblers or personal betting.

In this paper we develop machine learning models in order to predict outcomes of the English twenty over county cricket cup over the years 2009-2014. We used a multi-step approach to analyze the data that produced more than 500 features. We first team data only and then team paired with player data.

---

[1] Cricinfo Feature: http://www.espncricinfo.com/ci-icc/content/story/710723.html
[2] Cricinfo Feature: http://www.espncricinfo.com/pakistan/content/story/535250.html

Pearson correlation, mutual information the chi-square test, and recursive feature elimination were used for feature selection. The selected features were used as inputs to four different classification algorithms: naive Bayes, logistic regression, random forests and gradient boosted decision trees. Principle component analysis was also assessed as a way to improve the performance of the models.

The best model outperformed the odds provided by the bookmakers. The characteristics and performance of each model and directions for future research are discussed.

# 2 Data and Methods
## 2.1 Overview
Figure 1 below shows a flowchart of data collection and analysis.

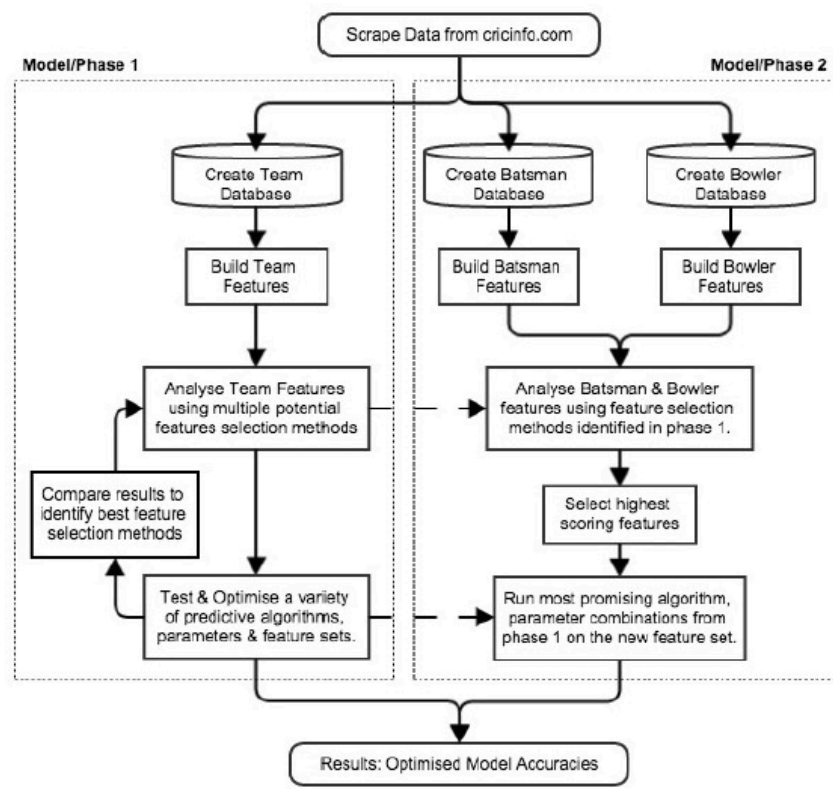

*Figure 1. Methodology flowchart*

The first major decision was to select the annual English county T20 cricket competition as the case study for this work. There were a number of reasons for this decision. International fixtures in cricket are potentially difficult to predict as all the nations do not play each other regularly enough to provide sufficiently deep training data, (Brooks, Faff, & Sokulsky, 2002). Furthermore, T20 is a major growth-area within cricket currently and attracts significant media interest globally. Also, the competition format has stayed roughly constant over the last few decade, unlike the longer one-day competition played at an English county level[3]. Finally, it is relatively new as a format and competition and consequently the coaches and players alike should be more engaged to learn in this area.

---
[3] Prior to 1999 there was only a 50-over competition. In 1999 a 40 competition was added. In 2010 both competitions were combined into one 40-over competition, which then became 50-overs in 2014.

The competition is played by the 18 counties with official first-class cricket status. The counties are split into regional groups which play each other at least once in a round robin fashion. The top few teams in each group go through to the quarter finals with the highest finishers playing at home. The semi-finals and finals are then held on one day at a predetermined venue. Over the years the format has varied between two groups of 9 and 3 groups of 6. The number of games per group has also varied, with teams playing each other both home and away in some years, only once in others and sometimes a combination in between (i.e. 3 of 5 teams twice and the other two only once).

Group games do not necessarily end with one side winning. They can either have 'no result' if weather prevents the game from being finished or be 'tied' if both teams end on the same score. In the knock out rounds, 'no result' games are replayed on a reserve day and 'tied' matches are decided by 'bowl-out', ensuring that there is always a winner.

Obviously games where there is 'no result' are not suitable for this work and so they were excluded from the data set. Games that take place on finals day were also excluded as they are the only 3 games in each season where neither of the teams has the home advantage and therefore would probably need to be modelled separately. 'Tied' matches are more complex but the decision was made to exclude them as well as they only occur in around 1% of instances and would therefore need rare event techniques, e.g. (King & Zeng, 2001), and would never be predicted over a win or loss anyway. Table 1 shows the number of games played in each season since 2003, how many were excluded and how many remained in the final dataset.

*Table 1. Historic English T20 matches by year*

| Year | # Games | Exclude: Finals Day | Exclude: Tied Matches | Exclude: No Results | Include |
|---|---|---|---|---|---|
| 2003 | 48 | 3 | - | - | 45 |
| 2004 | 52 | 3 | - | 4 | 45 |
| 2005 | 70 | 3 | - | 11 | 56 |
| 2006 | 70 | 3 | 1 | 2 | 64 |
| 2007 | 70 | 3 | 1 | 20 | 46 |
| 2008 | 97 | 3 | 3 | 10 | 81 |
| 2009 | 97 | 3 | - | 3 | 91 |
| 2010 | 151 | 3 | 3 | 5 | 140 |
| 2011 | 151 | 3 | 2 | 23 | 123 |
| 2012 | 97 | 3 | 1 | 20 | 73 |
| 2013 | 97 | 3 | 2 | 1 | 91 |
| 2014 | 133 | 3 | 1 | 12 | 117 |
| **Total** | **1,113** | **36** | **14** | **111** | **972** |

Historical match and player data for the English County twenty over (T20) competition was scraped from the archive section of cricket fan site cricinfo.com[4] using a range of web crawlers written in Python. This raw data was then cleaned and combined into three datasets for matches/teams, batsmen and bowlers.

---
[4] cricinfo.com data archive: http://www.espncricinfo.com/ci/engine/series/index.htm

From these datasets, features sets are constructed that form the input for two-different model classes:

**Model 1**: **Team Data Only**
Statistical features are formed that represent the 'form' or skill level of a particular team. These features are formed within a clear hierarchy where level 1 features are basic performance statistics, level 2 features are combinations of 2 level 1 features and level 3 features are combinations of 2 level 2 features. 31 features were formed in all.

**Model 2**: **Player & Team Data**
Further features were formed by calculating individual stats for each member of a team for both batting and bowling. Again features fell into clear hierarchies of levels 1, 2 & 3, however this time there were more than 500 features. This feature space was narrowed down using methods that had been shown to work well during the first modelling phase. The optimal algorithms from phase 1 were retrained on this new data set to see if further improvements were possible.

2.2   Feature selection

Chi-square, mutual information and Pearson correlation scores were used as feature selection methods. It became clear from the initial feature analysis that it was important to study these selection metrics on a year by year basis. To be useful in a predictive capacity, a feature has to be informative regarding the outcome in the same way each year, rather than just when the dataset is viewed as a whole. This is because an average value across all 12 years can hide some surprising fluctuations that happen on a year to year basis.

Figure 2 shows both a decreasing trend in home advantage over the years and also shows anomalies in 2004, 2012 and 2014 where the advantage appears to lie with the visiting team; Figure 3 plots the correlation scores of 'net run rate difference' against 'home team win percentages' by year and as expected there is a strong overall correlation of 0.99, however when viewed on a year by year basis there are again clear anomalies within 2004 and 2010. Obviously, anomalies of this type or concept drift will cause significant issues when predicting affected seasons based on previous season data.

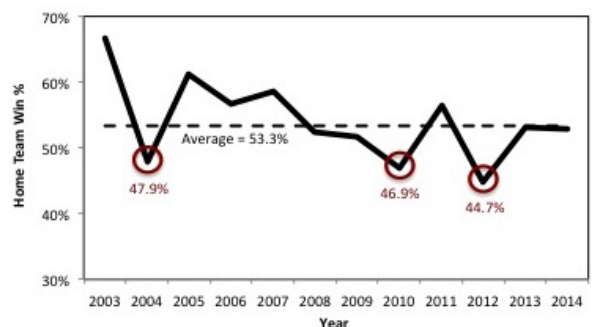

*Figure 2. Home advantage variation by season, highlighting anomalous years*

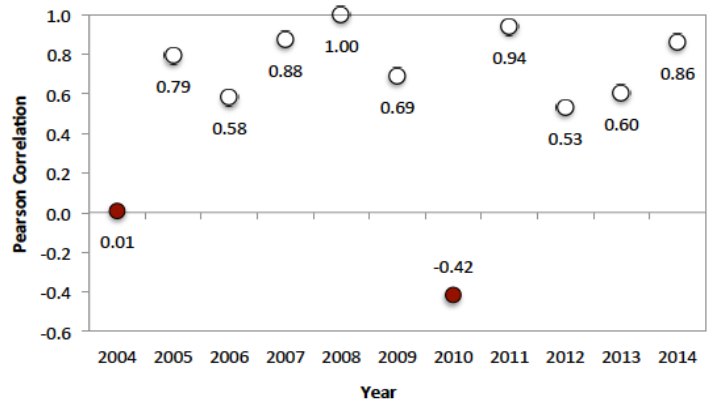
*Figure 3. Correlation of net run rate with win percentage by year*

## 2.3 Predictive modelling and assessment

The algorithms used for predictive modelling were: Naïve Bayes, logistic regression, random forests and gradient boosted trees. It was also investigated whether implementing principal component analysis (PCA) prior to learning would have a beneficial affect. The 6 seasons from 2009 to 2014 were chosen to be test years.

The performance of each predictive algorithm was assessed using the previous year's data as the training data and the year in question as the test set. Each model was tested over 6 seasons from 2009-2014. Classification accuracy (correct predictions / total predictions) was taken as the performance metric. Calculated both at an annual level and as a combined average.

In order to provide context for the prediction accuracy achieved, a benchmark was created using historic betting odds data. This benchmark was calculated as the % of times the bookies pre-match favorite wins.

## 2.4 Model 1: Team data

Nine different statistics were used as a starting point for creating the team level features and they are outlined in Table 2. They can be calculated for both the home and away teams, giving 18 base features. Whilst the majority are recognized performance metrics, batting wicket rate is not usually considered but is included here for completeness.

*Table 2. Team features*

| Team Feature | Description |
|---|---|
| Win Percentages | For previous $x$ games: Games won / $x$. |
| Batting Run Rate | Runs scored in previous $x$ games / Overs batted. |
| Bowling Economy Rate | Runs conceded in previous $x$ games / Overs bowled. |
| Batting Average | Runs scored in previous $x$ games / Wickets lost. |
| Bowling Average | Runs conceded in previous $x$ games / Wickets taken. |
| Batting Wicket Rate | Wickets lost in previous $x$ games / Balls batted. |
| Bowling Strike Rate | Wickets taken in previous $x$ games / Balls bowled. |
| Batting Index | Batting Run Rate x Batting Average. |
| Bowling Index | Bowling Economy Rate x Bowling Average. |

The batting and bowling performance indices are based on the individual player indices proposed by Sky shown in Figure 4 but applied at a team level. The batting index is a product of a team's run rate and average rather than the sum, based on comments from (Kumar, 2014). The corresponding bowling index was then calculated in order to maintain parity of batting and bowling features.

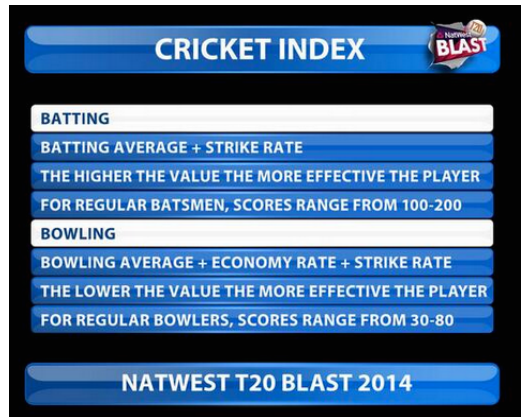

*Figure 4. Sky batting & bowling indices*

From these base or level 1 features it is possible to calculate more complex features in the following ways:

1. **Net Features**: These are the batting stats for a given team minus the bowling stats for the same team. For example, net run rate is the batting run rate minus the bowling economy rate. This obviously does not apply to the win percentages. There are 4 of these features for each team, i.e. 8 in total and they will be referred to as net or level 2 features throughout the remainder of the work.

2. **Difference Features**: These are the differences in both the win percentages and the net stats of the two teams. For example, Net Run Rate Difference is the Home Team's Net Run Rate - the Away Team's Net Run Rate. This gives 5 new difference or level 3 features.

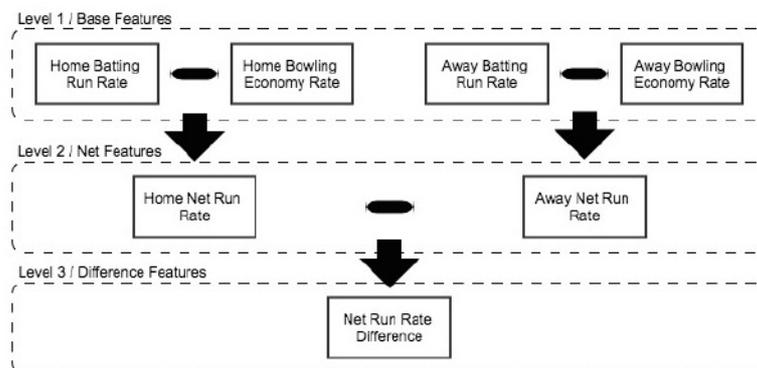

*Figure 5. Illustration of feature structure: Run Rate example.*

These 31 features (18 x level 1, 8 x level 2 & 5 x level 3) were calculated for each match in the dataset, using the previous games played by each team to do so.

The 6 main feature sets were as follows:
1. **14 level 1 features:** 2 x Win Percentages, 2 x Batting Run Rate, 2 x Bowling Economy Rate, 4 x Averages, 2 x Batting Wicket Rate, 2 x Bowling Strike Rate.
2. **18 level 1 features:** Same as 1 but + 2 x Batting Index & 2 x Bowling Index.
3. **8 level 2 features:** 2 x Win Percentages, 2 x net run rate, 2 x net average, 2 x net wicket rate.
4. **10 level 2 features:** Same as 3 but + net bowling and net batting indices.
5. **4 level 3 features:** Win difference, run rate difference, average difference, wicket rate difference.
6. **5 level 3 features:** Same as 5 but + index difference.

## 2.5 Model 2: Team & Player data

The focus of this second model was to use player related statistical features, firstly to increase overall classification accuracy and secondly to identify which of the 500 new features best help to do this.

**Level 1 Player Features**: As in phase 1, base or level 1 features for individual players were generated and then used to generate more complicated or higher level features in turn. These level 1 player features are given in Table 3 and Table 4 for batting and bowling respectively along with the method used to calculate them from raw data.

*Table 3. Batting features: Level 1*

| Batting Features | Description |
|---|---|
| Batting Average | Runs scored / number of times out. |
| Batting Strike Rate | Runs scored / balls faced. |

*Table 4. Bowling features: Level 1*

| Bowling Features | Description |
|---|---|
| Bowling Average | Runs conceded / wickets taken. |
| Bowling Economy Rate | Runs conceded / overs bowled. |
| Bowling Strike Rate | Balls bowled / wickets taken. |

As well as being calculated for individual players, these level 1 features were calculated for logical combinations of players such as the top 3 batsman or the top 5 bowlers. This significantly increased the number of potential level 1 features, the combinations explored for batting and bowling were as follows:

**Different Batsman Combinations** (28 in total): 11 individual batsman, top 2, top 3, top 4, top 5, top 6, top 7, top 8, top 9, top 10, all 11, numbers 5 to 8, 4 to 6, 7 to 9, 3 to 4, 5 to 6, 7 to 8, & 9 to 10.

**Bowling** (11 in total): 6 individual bowlers, top 2, top 3, top 4, top 5, all 6.

This therefore gave 89 base 1 features for each team (28 x 2 batting + 11 x 3 bowling), or 178 in total.

**Level 2 Player Features**: As before, level 2 features can be created by combining level 1 features. This was done in 1 of two general ways:

1. **Difference Method**: Subtracting the away feature from the corresponding home feature. For example, the average for the home side batsman 1 minus the average for the away side batsman 1. 28 level 2 features were created in this way.
2. **Combination Metrics**: Combine two or more level 1 features for a given player or set of players to create a new performance metric. i.e. Home Batsman 1 average + Home Batsman 1 Strike Rate. A number of ways of doing this were explored and they are outlined in Table 5 and Table 6. This combination of ideas is taken from Sky's performance index and (Kumar, 2014). 100 new level 2 features per team were calculated in this manner.

*Table 5. Combinations of level 1 features used to create level 2 batting features*

| Batting Features | Feature Type & Description |
|---|---|
| Average + Strike Rate | T20 batting performance index used by Sky. |
| Average x Strike Rate | Alternative to Sky batting index. |

*Table 6. Combinations of level 1 features used to create level 2 bowling features*

| Bowling Features | Feature Type & Description |
|---|---|
| Average + Economy Rate + Strike Rate | T20 bowling performance index used by Sky. |
| Average x Economy Rate x Strike Rate | Alternative 1 to Sky batting index. |
| Average + Economy Rate | Alternative 2 to Sky bowling index. |
| Average x Economy Rate | Alternative 3 to Sky bowling index. |

This is a further 228 level 2 features in total.

**Level 3 Player Features**: Finally, the level 2 features calculated as combination features can be compared between the home and away sides to calculate level 3 features. For example, 'Average x Economy Rate' for the home top 3 bowlers minus 'Average x Economy Rate' for the away top 3 bowlers. This gives rise to a further 100 level 3 features, and 506 features overall.

## 2.6 Complications due to new players:

Players who had not played in the competition before generated a couple of issues in building the player features. In general, these players fall into two different groups:

1. **Homegrown Talent**: British players who have come through county development set ups. They will likely be young and inexperienced and therefore perhaps less impactful.
2. **Overseas Players**: Each team is allowed two non-English/Welsh players and as such these are likely to be skilled and experienced T20 professionals even if they have not played in the English league previously.

The difference in performance of these two groups over their first games in the competition is significant (Figure 6). To capture this, average values were calculated on a cumulative basis from prior seasons, by batting position and by year for both groups separately. These average values were then used in place of real values for any batsman who had played fewer than 4 games.

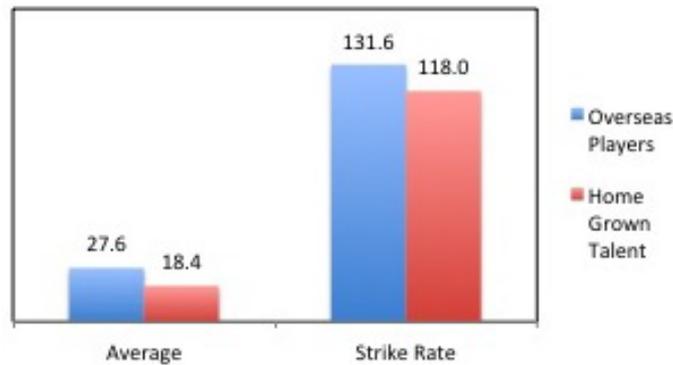

*Figure 6. New player performance metrics: Overseas professionals vs. Homegrown talent.*

Bowlers present a more difficult problem as before the game starts it might not be clear which of the 11 players will bowl, how many overs each will bowl and in what order they will bowl them. To circumvent this the player's values for each statistic were ranked in terms of their previous performances and if fewer than 6 players in a team have bowled a sufficient number of overs then average values were used instead. In this case it was not possible to use the overseas vs. home grown talent classification, given that prior to a game it is not known who will bowl or not.

## 3 Results

### 3.1 Model 1: Team data

#### 3.1.1 Feature selection

Chi-square, mutual information and Pearson correlation scores were calculated for bucketed versions of the 31 team features.

The scores presented here are represented by average values calculated across years 2006-2014. The variance of these annual scores is also presented to reflect the level of annual fluctuation. Good features should score highly and consistently in each year.

An initial observation is that the relative attractiveness of the features can depend on which statistical measure is considered. An example of this for the 5 level 3 team features is shown in Figure 7. As Pearson correlations can be negative, it is the magnitude that is displayed here for ease of understanding. With regards to the chi-square p-values it is worth noting that low scores are preferable.

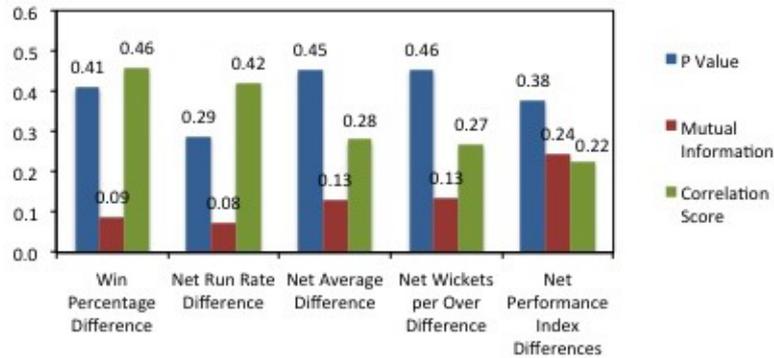

*Figure 7. Mutual Information, Chi-Square & Pearson Correlation scores for the 5 level 3 team features*

Features such as the 'Net Performance Index Difference' which scores relatively well from a p-value and mutual information perspective but not in terms of correlation, will most likely have a non-linear relationship with the outcome. Other features that show correlation but score less well in the other two such as run rates and wicket rates will likely have linear relationships with the outcome but have less spread.

Despite some of the differences, all 3 feature selection metrics seem to suggest that the higher the hierarchy level, the more informative the feature. Figure 8 demonstrates this for the performance index (PI) features and the mutual information scores.

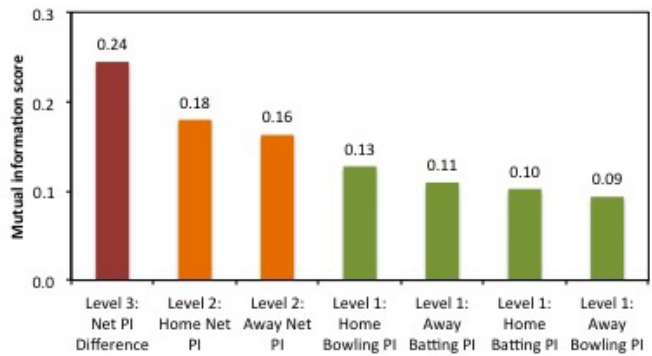

*Figure 8. Team performance index mutual information scores across hierarchy levels.*

As well as scoring highly on an average basis, the higher level features appeared to show less variance than the level 1 features. An example is illustrated by Figure 9 and Table 7 for the win percentage features. The level 2 'win difference' feature is not as correlated with the outcome as the level 1 'home team win percentage' feature. Despite this however, the win difference might still be the more attractive feature as the variance is significantly lower across the different seasons of the competition, meaning that the relationship is more consistent from year to year.

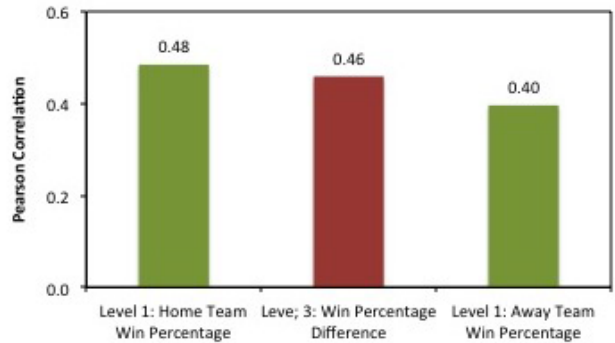
*Figure 9. Win percentage Pearson correlation scores*

*Table 7. Variance of the team win percentage Pearson correlation scores by year*

| Level 1: Home Team Win Percentage | Level 3: Win Difference | Level 1: Away Team Win Percentage |
|---|---|---|
| 0.181 | 0.071 | 0.067 |

### 3.1.2 Predictive Modelling

Year by year comparisons of the four optimized learners demonstrate clearly how some years appear to be harder to predict than others. With 2010 in particular and 2013 to a lesser degree appearing anomalous. It is worth noting that a number of learners performed significantly better within these years, but as a result the performance across the remaining years suffered and the average accuracy was low.

Figure 10 below shows the average accuracy per season per classifier, while Table 8 shows the average accuracy for each algorithm.

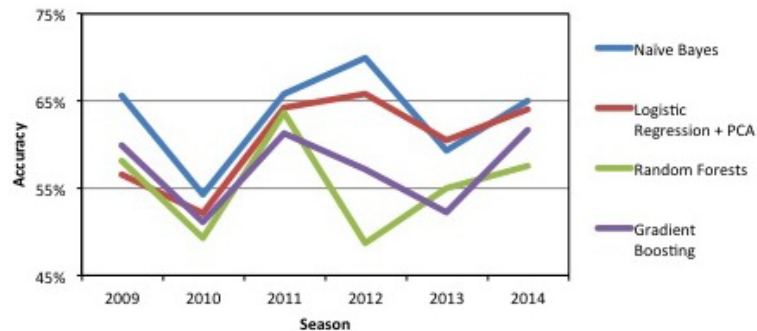
*Figure 10. Model 1 summary results by algorithm and test year*

*Table 8. Model 1 average accuracy by classification algorithm*

| Naive Bayes | Logistic Regression + PCA | Random Forests | Gradient Boosting |
|---|---|---|---|
| 62.4% | 60.1% | 55.6% | 57.2% |

## 3.2 Model 2: Team and Player Data and benchmark

The focus of this second model was to use player related statistical features, firstly to increase overall classification accuracy and secondly to identify which of the 500 new features best help to do this.

### 3.2.1 Feature selection

As before, general trends seen in the results of the feature selection process. Once again the higher level features display the strongest results, however to a lesser degree than for the team features. From phase 1 it was observed that the Pearson correlation score best represented features that went on to perform well in a predictive setting and therefore that is was is considered here within phase 2.

**Batting Features**: There is a consistent and clear order in terms of which batting metrics produce the most extreme Pearson correlation scores against the 'Home Win Percentage', which is:

Strike Rate + Average (Level 2) > Strike Rate x Average (Level 2) > Strike Rate (Level 1) > Average (Level 1)

This is illustrated in Figure 11, using the average value of the top 8 batsman in each team as the example. A second important trend can be seen in Figure 11 which is:

Away Team Batting Features (Level 1 or 2)> Batting Difference Method Features (Level 2 or 3) > Home Team Batting Features (Level 1 or 2)

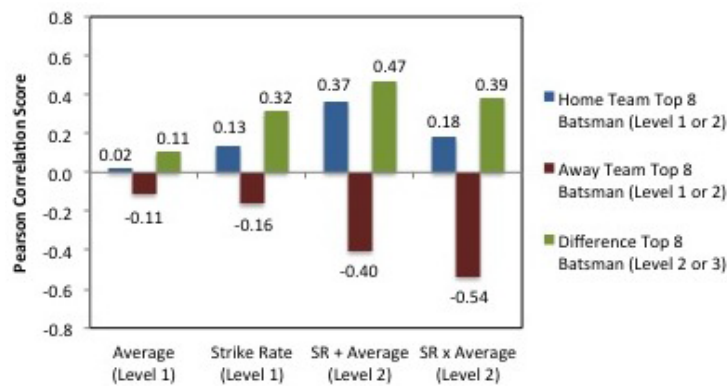

*Figure 11. Comparison of Pearson Correlation Scores for different batting statistics calculated using the top 8 batsman in each team*

**Bowling Features**: For the bowling metrics, it is less clear which show the most significant correlation with the 'Home Win Percentage'. In most cases the correlation scores are fairly similar across all seven metrics, as shown in Figure 12 for the top 3 bowlers in each team. However, the following two observations can be made:

1) Bowling economy rates generally show slightly higher correlation than bowling averages and strike rates.
2) The sky bowling index (Average + Economy Rate + Strike Rate) generally show higher correlation than the other 3 combination metrics.

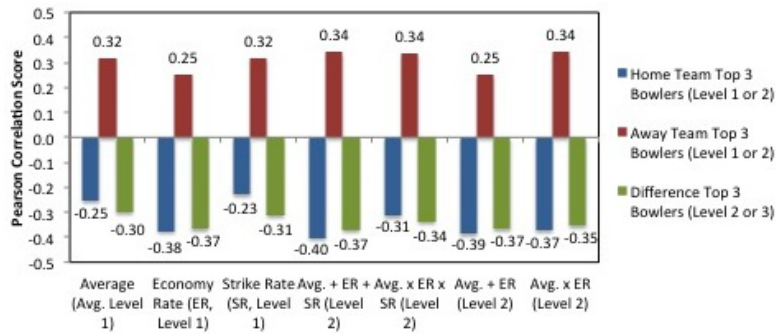

*Figure 12. Comparison of Pearson Correlation Scores for different bowling statistics calculated using the top 3 bowlers in each team*

Unlike the for the batting metrics, home and away metrics show trends of similar magnitudes. In general, the difference metrics the at least as correlated as for home and away as can be seen in Figure 12 and Figure 13.

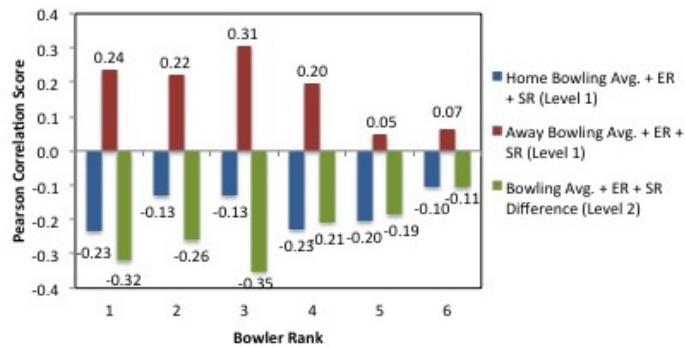

*Figure 13. Pearson Correlation score for individual bowler's sum bowling index by bowling rank.*

### 3.2.2 Predictive modelling

**Naïve Bayes**

The 50 or so, highest scoring features from the previous section (in terms of Pearson correlation), were combined with the two optimal team features discovered in model 1. This feature set was then used as the input to a naive Bayes model used in conjunction with recursive feature elimination. This led to a new optimal feature set of 4 features that displayed improved accuracy when compared to the team only model. This improvement can be seen on a year by year basis in Figure 14 and in summarized form over the 6 years in Table 9.

Also of interest was the optimal 5 feature set discovered in this way. Although slightly less accurate overall, it achieved relatively high accuracy in 2010 and 2013 which have been difficult to model throughout this work. This improvement does come at the cost of reduced accuracy in 2009 and 2011 however.

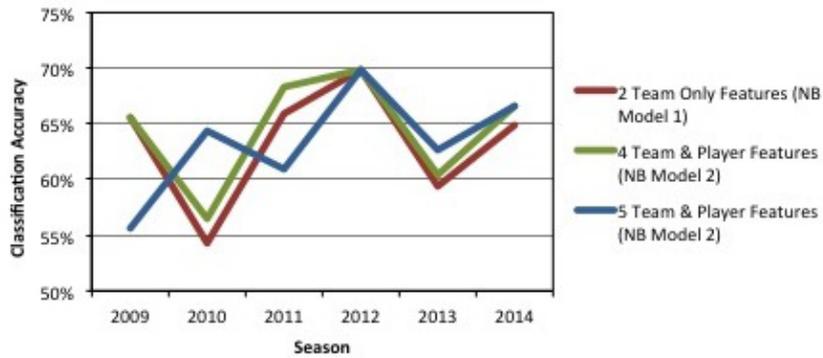

*Figure 14. Naive Bayes improvements through the addition of player data.*

*Table 9. Naïve Bayes model 2 classification accuracies*

| Feature Space | Average Accuracy (2009-2014) | Average Accuracy (2010-2014) |
|---|---|---|
| 2 Team Features (Model 1) | 62.6% | 62.1% |
| 4 Team & Player Features | 64.0% | 63.8% |
| 5 Team & Player Features | 53.2% | 64.5% |

**Random Forests**

In phase 1, the optimum data set found in conjunction with random forests was using the 10 level 2 or net features. Based on the Pearson scores from the previous section, two further data sets were constructed, using either 20 further level 2 features were added.

Accuracy improved from 55.6% to 58.1%, however performance remained sporadic when viewed on a year by year basis. This is shown in Figure 15.

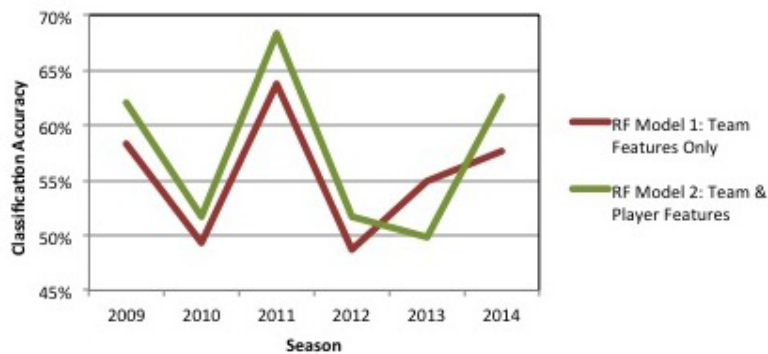

*Figure 15. Random Forests improvements through the addition of player data*

## 3.3 Comparison against odds

As mentioned previously, an accuracy benchmark was calculated from the betting industry to provide some context for the results achieved as part of this project. Historic odds data was found from an online betting portal for matches as far back as 2009, but the data was only found to be reliable and compete from 2011 onwards. The accuracy benchmark was calculated as the percentage of times the bookies favorite went on to win a match. The comparison vs. the optimal naive Bayes predictor is displayed in Figure 16.

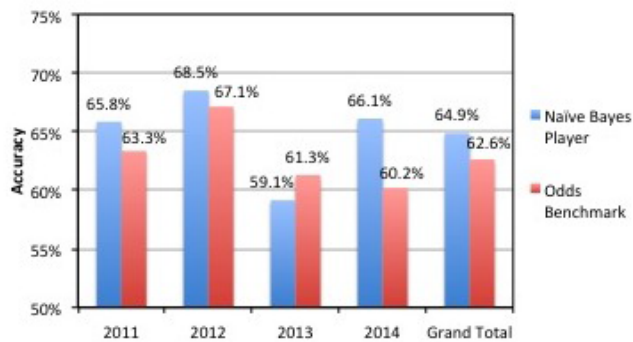

*Figure 16. Optimized naive Bayes team & player model vs. the betting benchmark*

In many similar academic studies, e.g. (Warner, 2010), at this point it is usual to find that any accuracy level achievable with machine learning methods is lower than that found in the gambling industry. The relatively consistent over performance by the naive Bayes learner displayed in Figure 16 is therefore surprising and represents a potential financial opportunity.

Studying the accuracy fluctuations across a season shows reveals that although unstable at the beginning of a season, the performance of the optimized classifier settles down to an average value relatively quickly. Often, as shown in Figure 17 for 2014, this average accuracy is consistently above the odds benchmark.

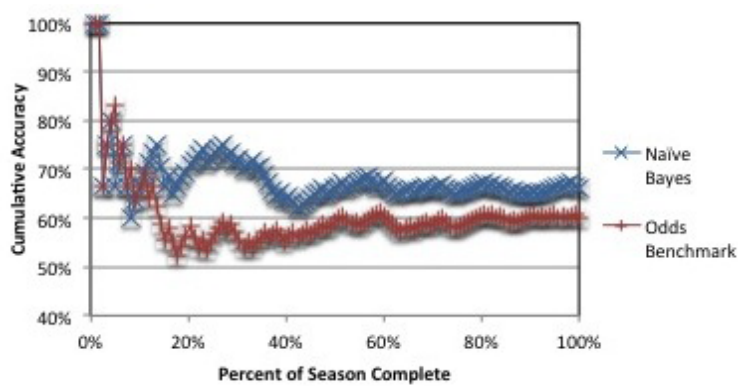

*Figure 17. Cumulative accuracy variation across the season 2014. Naive Bayes vs. the betting industry benchmark.*

## 4 Discussion and conclusions

It is possible to predict the winner of English county twenty twenty cricket games in almost two thirds of instances. This is an improvement upon levels present in the gambling industry today and implies a potential financial opportunity. However, the overall level of accuracy is lower than that observed in many other sports, (Purucker, 1996; Pardee, 1999; Zimmerman, Moorthy, & Shi, 2013) and undergoes more significant fluctuations on a yearly basis, (Zimmerman, Moorthy, & Shi, 2013). This suggests that twenty twenty cricket has a relatively high level of instability or randomness within it, which should come as no surprise to those familiar with the game.

Of the various methods tried, the most effective classification method was a simple naive Bayes learner combined with significant data preprocessing, feature selection and complex hierarchical features.

Pearson correlation scores appear more effective than both mutual information and chi-square scores when selecting statistical features for this problem. This is likely to be because the majority of relationships between the features and the outcome are linear, as Pearson correlation is designed to identify linear relationships. This seems reasonable given the scenario of statistics reflecting player or team ability being compared against a team's chances of winning.

With regards to future work, further avenues could be explored either in the form of new algorithms (as mentioned in the previous section) or in the form of new features. This work has focused entirely on using team or player performance based statistics as features. Whereas previous investigations (Bandulasiri, 2008; Kaluarachchi & Varde, 2010) use predominantly categorical features, such as 'home team', 'who won the toss' or 'venue' for the same task. Combining these feature types could perhaps yield performance gains, however algorithm choice would be limited to those capable of dealing with both categorical and numerical features concurrently. In addition, external data such as the weather or social media comments could be mined to provide even more features for further experimentation.